\pdfoutput=1

\documentclass[11pt]{article}

\usepackage[final]{acl}

\usepackage{times}
\usepackage{latexsym}

\usepackage{tabularx} 
\usepackage{makecell} 
\usepackage{booktabs}
\usepackage{color, colortbl}
\usepackage{amsmath}
\usepackage{tcolorbox}
\newtcbox{\mybox}[1][red]
  {on line, arc = 0pt, outer arc = 0pt,
    colback = #1!10!white, colframe = #1!50!black,
    boxsep = 0pt, left = 1pt, right = 1pt, top = 2pt, bottom = 2pt,
    boxrule = 0pt, bottomrule = 1pt, toprule = 1pt}
\usepackage[T1]{fontenc}

\usepackage[utf8]{inputenc}

\usepackage{microtype}

\title{UBARv2: Towards Mitigating Exposure Bias in Task-Oriented Dialogs}

\author{Yunyi Yang, Hong Ding, Qingyi Liu, Xiaojun Quan\textsuperscript{*} \\
    Sun Yat-sen University \\
    \texttt{\{yangyy37, dingh35, liuqy95\}@mail2.sysu.edu.cn, quanxj3@mail.sysu.edu.cn} \\
}

\begin{document}
\maketitle
\begin{abstract}
This paper studies the exposure bias problem in task-oriented dialog systems, where the model's generated content over multiple turns drives the dialog context away from the ground-truth distribution at training time, introducing error propagation and damaging the robustness of the TOD system.
To bridge the gap between training and inference for multi-turn task-oriented dialogs, we propose session-level sampling which explicitly exposes the model to sampled generated content of dialog context during training. Additionally, we employ a dropout-based consistency regularization with the masking strategy R-Mask to further improve the robustness and performance of the model. The proposed UBARv2 achieves state-of-the-art performance on the standardized evaluation benchmark MultiWOZ and extensive experiments show the effectiveness of the proposed methods. 
\end{abstract}

\section{Introduction}
Task-oriented dialog (TOD) systems assist users with various tasks via natural language conversations.
The traditional task-oriented dialog systems follow a pipeline approach which consists of several consecutive modules. First, a dialog state tracker (DST) is to estimate the belief state from the user utterance. The belief state is then used to query a task-related database (DB) like the number of entities that match the user's goal. Subsequently, a dialog policy learning module is applied to determine the next system act, followed by a natural language generation (NLG) module that converts the system action to a natural language response.

Recently, task-oriented dialog systems have achieved promising results by leveraging pre-trained language models \cite{radford2018improving} for end-to-end modeling in a unified way \cite{ham2020end,hosseini-asl2020a,peng2020soloist,yang2021ubar}. These works cast task-oriented dialogs as a unified language generation task and fine-tune models with the language modeling objective.
Particularly, UBAR \cite{yang2021ubar} models task-oriented dialogs on a dialog session level, which is trained on the sequence of the entire dialog session composed of user utterance, belief state, database result, system act, and system response of every dialog turn. During inference, the dialog context uses the generated content rather than the ground-truth annotations.
The successive works MTTOD \cite{lee2021improving}, PPTOD \cite{su2021multi} and GALAXY \cite{he2021galaxy} all follow such session-level modeling as the fundamental design when developing their methods. They achieve increasingly competitive performances via multi-task learning and large-scale in-domain pre-training. 

Despite the effectiveness of session-level modeling, bringing in generated content at inference time inevitably introduces a gap between training and inference. Since the distribution of the ground-truth annotations at training time is different from the distribution of the model predictions at inference time. If a mistake has occurred in the dialog context, there can be error propagation which causes the model generation to continue to deviate from the optimal distribution. This problem is often referred to as exposure bias for auto-regressive models. 
In the case of TOD systems, the exposure bias problem can take place across multiple modules over multiple dialog turns. 
For example, a TOD system is asking about the food style of the requested restaurant, but the user replies with the price range. 
Though the model might be able to update the belief state correctly, it could be confused when generating the system action and response of the next turn, given that the training data is coherent and consistent while the model has not seen such off-the-mark answers during training. 
What's more, being exposed to an unfamiliar situation where the dialog context contains low quality and erroneous generated content is detrimental to the model's performance and robustness.

In an attempt to mitigate the exposure bias problem that exhibits in task-oriented dialog systems, we follow the session-level modeling of UBAR and propose a learning framework UBARv2 that explicitly exposes the model to heterogeneous data at training time. 
Specifically, we explore the sampling strategy for constructing the session-level training sequence and perform mixed training from both the distribution of the annotation and the distribution of the model prediction. 
In the initial stage of training, the ground truth sequence is learned to help the model converge quickly, and then the content generated by the model is sampled on turn level with a certain probability for mixed training. We further employ a dropout-based consistency regularization with a masking strategy named R-Mask, which carries out the forward pass twice for the partially masked session-level sequence and optimize the bi-directional KL divergence loss of the two distributions, helping to bridge the gap between training and inference.

We conduct experiments on the MultiWOZ dataset \cite{budzianowski2018multiwoz} and use the standardized evaluation of Context-to-Response generation \cite{nekvinda2021shades} to compare UBARv2 with UBAR and other competitive baselines. 
UBARv2 greatly improves its predecessor UBAR and outperforms other state-of-the-art baselines in the end-to-end modeling setting.
We perform a thorough analysis to verify the effectiveness of the proposed method. 
In summary, our main contributions are as follows:
\begin{itemize}
    \item To the best of our knowledge, we are the first to study the exposure bias problem in task-oriented dialogs.
    \item We propose two effective strategies bridging the gap between training and inference  of task-oriented dialog systems.
    \item Experiments show that UBARv2 achieves state-of-the-art performance on the standardized MultiWOZ benchmark. 
\end{itemize}



\section{Methodology}
In this section, we introduce the session-level modeling as the building block of UBARv2 and describe the proposed session-level scheduled sampling strategy and consistency regularization method R-Mask. Figure \ref{fig:model} is an overview of UBARv2.
\subsection{Session-Level Modeling}
Session-level modeling is first introduced by \citet{yang2021ubar} and adopted by numerous successive methods \cite{lee2021improving,su2021multi,he2021galaxy}. Two key factors of session-level modeling contribute to the effectiveness of a task-oriented dialog system: the incorporation of intermediate information such as belief state and system action into dialog context, and using all generated content in the dialog context.

\begin{figure*}[h]
    \centering
    \includegraphics[width=1\textwidth]{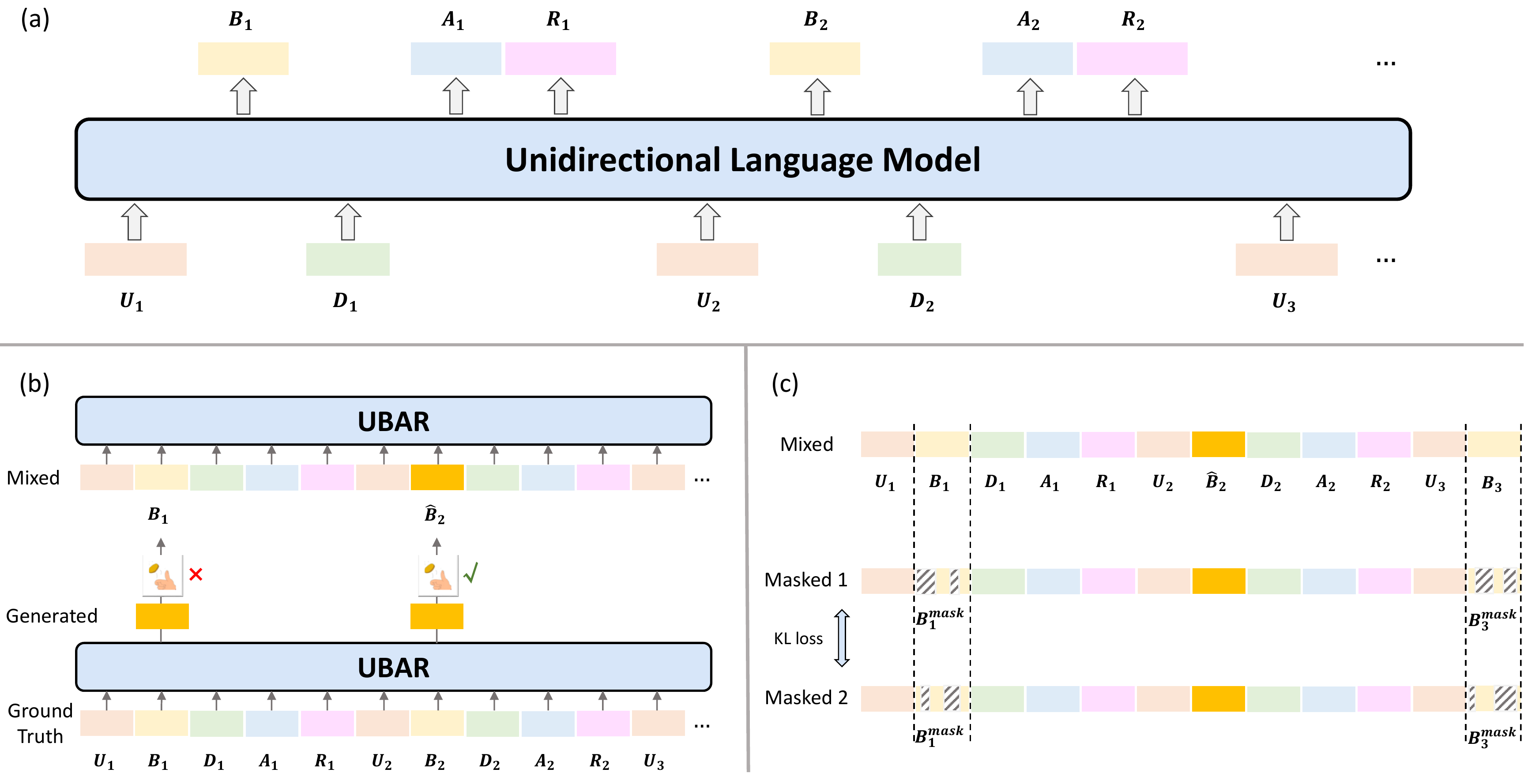}
    \caption{An overview of UBARv2, with session-level modeling, session-level sampling and R-Mask.}
    \label{fig:model}
\end{figure*}

As illustrated in Figure \ref{fig:model} (a), given a dialog session composed of multiple turns, the session-level modeling operates the process of a task-oriented dialog session as follows: In the first turn $\tau=0$, the user puts in user utterance $U_0$, based on $U_0$, the model generates a belief state $\hat{B}_0$. 
The belief state is applied to query a database to retrieve database search result $\hat{D}_0$, which is the matched number of entities that satisfy the constraint inflicted by the belief state.
Based on $\{U_0, \hat{B}_0, \hat{D}_0\}$, the model generates system action $\hat{A}_0$ and system response $\hat{R}_0$, accomplishing the interaction of the first turn. 
As the dialog proceeds to turn $\tau$, the model generates $\hat{B}_\tau$, $\hat{A}_\tau$ and $\hat{R}_\tau$ based on context of user utterances and all previous generated outputs $\{U_0, \hat{B}_0, \hat{D}_0, \hat{A}_0, \hat{R}_0, ... , U_{\tau-1}, \hat{B}_{\tau-1}, \hat{D}_{\tau-1}, \hat{A}_{\tau-1}\\, \hat{R}_{\tau-1}, U_{\tau}\}$, eventually concluding the entire dialog session.

The model can be trained with language modeling objective \cite{bengio2003a} for GPT-2-based architecture. The idea of session-level modeling also applies for Seq2Seq architectures \cite{su2021multi,lee2021improving} or unified language model \cite{he2021galaxy}.

\subsection{Session-Level Sampling}
To bridge the gap between training and inference, we can draw inspiration from the domain of neural machine translation, where scheduled sampling is employed such that the input to the decoder at time step $t$ is chosen randomly between the ground-truth word and the model's prediction \cite{bengio2015scheduled,zhang2019bridging}. 

Instead of simply considering the word-level exposure bias of autoregressive generation by sampling context words, this work focuses on addressing the discrepancy across multiple turns in a dialog session. Therefore, we propose to sample content in a turn-wise and modular-wise manner and construct session-level sequence mixed with ground-truth and generated modular spans for training.
Specifically, as shown in Figure \ref{fig:model} (b), in every turn we can decide with a certain probability whether to sample a generated modular span, such as belief state, database result, system action and system response. Take belief state for example, which is the determined sampling target of our method. As the dialog proceeds to turn $\tau$, UBARv2 takes dialog context $\{U_0, B_0, D_0, A_0, R_0, ... , U_{\tau-1}, B_{\tau-1}, D_{\tau-1}, A_{\tau-1}, \\R_{\tau-1}, U_{\tau}\}$ and generates $\hat{B}_\tau$. 
We choose with probability $\epsilon$ to sample the generated belief state span $\hat{B}_\tau$ and with probability $(1-\epsilon)$ to sample the ground-truth span $B_\tau$.
Performing sampling with chance every turn results in a full dialog session-level training sequence of $M$ turns: $\{U_0, \hat{B}_0, D_0, A_0, R_0, ... ,$ $U_M, \hat{B}_M, D_M, A_M, R_M\}$, where $\hat{B}_\tau$ is the generated content.

At the early stage (Stage 1) of learning, the model is trained with ground-truth sequences so that UBARv2 can effectively learn task-oriented dialogs. At the late stage (Stage 2), the model employs mixed training with sampling rate $\epsilon$ and exposes itself to the inference setting, learning to deal with inconsistent and incoherent dialogs. The objective is to minimize the negative log-likelihood of the session-level sampled sequence $\hat{\mathbf{x}}=\{\hat{x}_0,\hat{x}_1, ..., \hat{x}_T\}$:
\begin{equation}
    \mathcal{L}_{\widehat{NLL}} = - \sum_{t=1}^{T} \log P_{\theta}\left(\hat{x}_i \mid \hat{\mathbf{x}}_{<t}\right)
\end{equation}

It is important to note that the model is trained on the sampled sequence instead of always trained on the ground-truth tokens based on the last sampled word like previous methods \cite{bengio2015scheduled,zhang2019bridging}.
What's more, We fix the sampling rate $\epsilon$ at Stage 2 instead of using scheduled rate or with decay for simplicity, while we focus more on the strategy such as which component of the dialog context to sample.

\subsection{R-Mask}
Inspired by R-Drop \cite{wu2021r}, which attempts to make models with dropout \cite{srivastava2014dropout} be more consistent during training and inference, we explore dropout-based consistency training in helping mitigate the exposure bias problem in task-oriented dialogs. Other than explicitly making the model learn its sampled generated content, we hope such consistent regularization could expose the model to more non-ground-truth data, eventually reducing the gap. 

There are two scenarios applying consistency regularization to the generation task of TOD systems: Stage 1 training and Stage 2 training. At the Stage 1, which is the early stage when training on the ground-truth sequences $\mathbf{X}=\{x_0,x_1, ..., x_T\}$, the model goes through the forward pass twice and acquires two distinct distributions of the same sequence from the randomness in the model. The objective is to minimize the bidirectional KL divergence between the two distributions:
\begin{equation}
\begin{aligned}
\mathcal{L}_{KL} = \sum_{t=1}^{T} \frac{1}{2}  [D_{KL}\left(P_{\theta^{1}}\left(x_{t} \mid \mathbf{x}_{<t}\right) \| P_{\theta^{2}}\left(x_{t} \mid \mathbf{x}_{<t}\right)\right) \\ + D_{KL}\left(P_{\theta^{2}}\left(x_{t} \mid \mathbf{x}_{<t}\right) \| P_{\theta^{1}}\left(x_{t} \mid \mathbf{x}_{<t}\right)\right) ]
\end{aligned}
\end{equation}

In essence, by adding a KL divergence regularization term, R-Drop increases the robustness to dropout and forces the model output to be consistent under different dropouts. 

Consistency training can introduce model-level regularization and data-level regularization which involves modification of the input data.
Therefore, at Stage 2, the late training stage, we employ an addition masking strategy R-Mask to the sampled sequence to obtain different distributions. As shown in figure \ref{fig:model} (c), we randomly replace certain elements in the sampled sequence with the special token ``[MASK]'' and with more variants for the regularization term:
\begin{equation}
\begin{aligned}
\mathcal{L}_{\widehat{KL}} = \sum_{t=1}^{T} \frac{1}{2}  [D_{KL}\left(P_{\theta^{1}}\left(\hat{x}_{t} \mid \hat{\mathbf{x}}_{<t}\right) \| P_{\theta^{2}}\left(\hat{x}_{t} \mid \hat{\mathbf{x}}_{<t}\right)\right) \\ + D_{KL}\left(P_{\theta^{2}}\left(\hat{x}_{t} \mid \hat{\mathbf{x}}_{<t}\right) \| P_{\theta^{1}}\left(\hat{x}_{t} \mid \hat{\mathbf{x}}_{<t}\right)\right) ]
\end{aligned}
\end{equation}

It is important to maintain the KL divergence throughout training at both stages so that the model can be trained properly. If we apply this consistency training midway, the KL divergence loss would be too large and severely hinder the language modeling optimization. The training loss is a combination of language modeling loss and the KL divergence loss with hyper-parameter $\alpha$ as regularization weight:
\begin{equation}
\begin{aligned}
\mathcal{L}_{\text{Stage1}} &= \mathcal{L}_{NLL} + \alpha \mathcal{L}_{KL} \\ 
\mathcal{L}_{\text{Stage2}} &= \mathcal{L}_{\widehat{NLL}} + \alpha \mathcal{L}_{\widehat{KL}}  
\end{aligned}
\end{equation}



\section{Experiments}
\begin{table*}[ht]
    \centering
    \setlength\tabcolsep{15pt}
    \renewcommand{\arraystretch}{1.2}
    \begin{tabular}{l | c c c c}
        \toprule
        Model & Inform & Success & BLEU & Comb \\
        \hline
        DAMD \cite{zhang2020task} & 57.9 & 47.6 & 16.4 & 84.8\\
        AuGPT \cite{kulhanek2021augpt} & 76.6 & 60.5 & 16.8 & 85.4 \\
        MinTL \cite{lin2020mintl} & 73.7 & 65.4 & 19.4 & 89.0 \\
        SOLOIST \cite{peng2020soloist} & 82.3 & 72.4 & 13.6 & 90.9\\
        UBAR \cite{yang2021ubar} & 83.7 & 70.3 & 17.6 & 94.4\\
        PPTOD \cite{su2021multi} & 83.1 & 72.7 & 18.2 & 96.1\\
        BORT \cite{sun2022bort} & 85.5 & 77.4 & 17.9 & 99.4 \\
        MTTOD \cite{lee2021improving} & 85.9 & 76.5 & 19.0 & 100.2\\
        GALAXY \cite{he2021galaxy} & 85.4 &	75.7 & \textbf{19.6} & 100.2 \\
        \hline
        UBARv1 & 82.1 & 69.7 & 17.9 & 93.8 \\
        UBARv1+SS & 83.9 & 71.0 & 17.6 & 95.0 \\
        UBARv1+R-Drop & 86.8 & 76.8 & 18.5 & 100.3 \\
        UBARv2 & \textbf{87.5} & \textbf{77.6} & 19.0 & \textbf{101.6} \\
        \bottomrule
    \end{tabular}
    \caption{
    Main results on MultiWOZ Evaluation End-to-end modeling.
    }
    \label{tab:v2_main_results}
\end{table*}

\subsection{Dataset and Evaluation Metrics}


MultiWOZ \cite{budzianowski2018multiwoz} is a human-human multi-turn task-oriented dialog dataset spanning multiple domains.
The dataset is divided into a training set containing 8438 dialogs, a verification set and a test set, both of which contain 1000 dialogs. Multiple versions of MultiWOZ \cite{budzianowski2018multiwoz,eric2019multiwoz,zang2020multiwoz} have been released as the benchmark \footnote{\url{https://github.com/budzianowski/multiwoz}} developing. For a fair comparison, this work conducts experiments and reports results based on the standardized evaluation scripts \footnote{\url{https://github.com/Tomiinek/MultiWOZ_Evaluation}} of MultiWOZ Evaluation \cite{nekvinda2021shades}.



We follow the automatic evaluation metrics to evaluate task completion and response quality: \textbf{Inform} measures whether the system provides an appropriate entity, \textbf{Success} measures whether the system answers all the requested attributes, and \textbf{BLEU} \cite{papineni2002bleu} is used to measure the fluency of the generated responses \cite{budzianowski2018multiwoz}. The BLEU score is calculated with references obtained from the MultiWOZ 2.2 span annotations \cite{nekvinda2021shades}.
A combined score: $(\textbf{Inform} + \textbf{Success}) \times 0.5 + \textbf{BLEU}$ is also reported as an overall quality measure suggested in \citet{mehri2019structured}.

\subsection{Implementation Details}

We initialize UBARv2 with DistilGPT2 \cite{sanh2019distilbert} and develop out method with HuggingFace's Transformers \cite{wolf2019huggingface}. Following \citet{zhang2020task}, the dataset is preprocessed using domain-adaptive delexicalization. 
We reimplement UBAR \cite{yang2021ubar} as UBARv1 and develop the two proposed method session-level sampling (SS) and R-Drop/R-Mask. Typically, UBARv1 and UBARv1+R-Drop and UBARv1+R-Drop are trained at Stage 1 for 60 to 75 epochs. UBARv1+SS is trained on top of UBARv1 at Stage 2 for 5 epochs. UBARv2, the final model is trained with R-Mask strategy at Stage 2 on top of UBARv1+R-Drop.
UBARv2 uses the strategy of sampling the belief state every turn for session-level sampling and masking the ground-truth belief state for R-Mask.
We select the model with the best performance on the validation set and evaluate it on the test set to get final results. The results in section \ref{sec:analysis} are mainly from the validation set. The batch size is 8, initial learning rate for AdamW is 1.5e-4, sampling rate $\epsilon$ for SS is 0.01, regularization weight $\alpha$ is 0.01 and the masking rate for R-Mask is 0.02. 
Code and models are included in the supplement and will be released \footnote{\url{https://github.com/dingdingtom/UBARv2}}.

\subsection{Baselines}
We compare UBARv2 with strong baselines on the benchmark as follows: DAMD \cite{zhang2020task}, MinTL \cite{lin2020mintl}, AuGPT \cite{kulhanek2021augpt}, SOLOIST \cite{peng2020soloist}, UBAR \cite{yang2021ubar}, GALAXY \cite{he2021galaxy}, PPTOD \cite{su2021multi}, BORT \cite{sun2022bort}, MTTOD \cite{lee2021improving}. 
UBARv2 is evaluated and compared in the end-to-end modeling setting of MultiWOZ, the system has to generate the belief state based on the context, query the database with that belief state, then generate the act and response. We also report the result of policy optimization which requires the model to generate system action and response based on ground-truth belief state in Appendix \ref{app:policy}.





\subsection{Overall Results}


As shown in Table \ref{tab:v2_main_results}, the proposed UBARv2 achieves the state-of-art performance in terms of inform rate, success rate, and combined score, surpassing the previous models MTTOD and GALAXY, raising the combined score by 1.4 points, which indicates that attempting to mitigate exposure bias in task-oriented dialogs can effectively improve the task completion ability of TOD systems. Note that UBARv2 does not require pre-training on supplementary data like SOLOIST, PPTOD, and GALAXY.
The results in the second group show the variations of UBARv2, which serves as an ablation study. For starters, UBARv1+SS scores higher inform rate and success rate and lifting the combined score by 1.2 over UBARv1, which demonstrates the effectiveness of session-level sampling. Introducing R-Drop to the training process can bring a significant performance boost, UBARv1+R-Drop jumps the combined score from 93.8 to 100.3, which shows the effectiveness of the dropout-based consistency regularization. Combining R-Mask and SS at Stage 2, UBARv2 shows that the two proposed methods are complementary to each other and can further push the state-of-the-art performance.

To examine the domain transfer ability of UBARv2 generalizing to unseen domains, we perform zero-shot and few-shot experiments in Appendix \ref{app:domain}.



\section{Analysis and Discussion}
\label{sec:analysis}
In this section, we provide a detailed discussion of the sampling strategy and the context used in constructing a mixed training sequence. We investigate how the Sampling rate $\epsilon$ and regularization weight $\alpha$ affect the model performance. We also discuss the R-Mask strategy and provide case study to show how can UBARv2 improve task completion and mitigate exposure bias.

\begin{figure*}[h]
    \centering
    \includegraphics[width=1.09\textwidth]{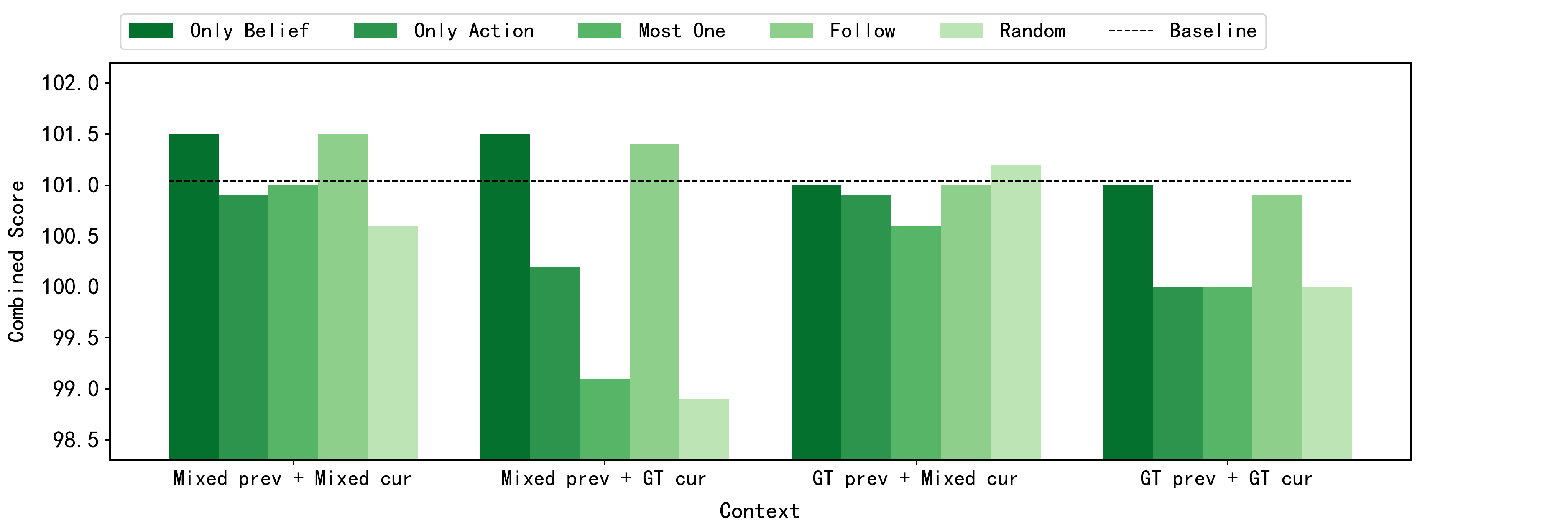}
    \caption{Sampling strategies with different sampling targets under different dialog contexts.}
    \label{fig:dependency}
\end{figure*}

\subsection{Sampling Strategy}
Sampling task-oriented dialogs and constructing mixed training sequences require a more fine-grained sampling strategy considering different TOD components such as belief state and system action with a dependent relationship. Additionally, we need to consider the attribute of the dialog context on which the sampling is conditioned.
First, we list five sampling strategies based on which components to sample in the current turn when constructing a sampled sequence:

\begin{itemize}
\item \textbf{Sampling only the belief state}: The annotated belief state will be replaced by the generated one in the corresponding position.

\item \textbf{Sampling only the system action}: The annotated system action and response will be replaced.

\item \textbf{Sampling at most one}: First determine whether to sample the belief state, and if so, use the annotated action. Otherwise, sample the action and response.

\item \textbf{System action follows the belief state}: Sample the belief state, action and response.

\item \textbf{Random Sampling}: The sampling of the action is independent of the one of the belief state.
\end{itemize}

Then, we divide the dialog context into the context of the previous turns and the context of the current turn, and consider whether they are (1) mixed, with some elements sampled, or (2) ground-truth, with all elements from the dataset.

As shown in Figure \ref{fig:dependency}, the dashed baseline is the validation score of UBARv1 + R-Drop and the histogram shows the score of UBARv1 + R-Drop after 5 epochs of mixed training with session-level sampling. For the sampling strategy, it can be seen that ``Sampling only the belief state'' and ``System action follows the belief state'' are more effective than the others. The effect of ``Sampling only the belief state'' is generally better than the one of ``Sampling only the system action'', which indicates that sampling the belief state is more meaningful than the action. 
For the context attributes, using the mixed context is always better than using the ground-truth one, generating content that is more fluent and more relevant to the previous context, which aligns with the intuition of session-level modeling. 
Note that when the sampling strategy is fixed as ``Sampling only the belief state'', the score is same for ``Mixed cur'' and ``GT cur''. This is because only the ground-truth user utterance is available when generating the belief state. 
Therefore, the sampling strategy for UBARv2 is ``Sampling only the belief state'' and ``Mixed context''.

\subsection{Sampling Rate}
\begin{figure}[h]
    \centering
    \includegraphics[width=0.52\textwidth]{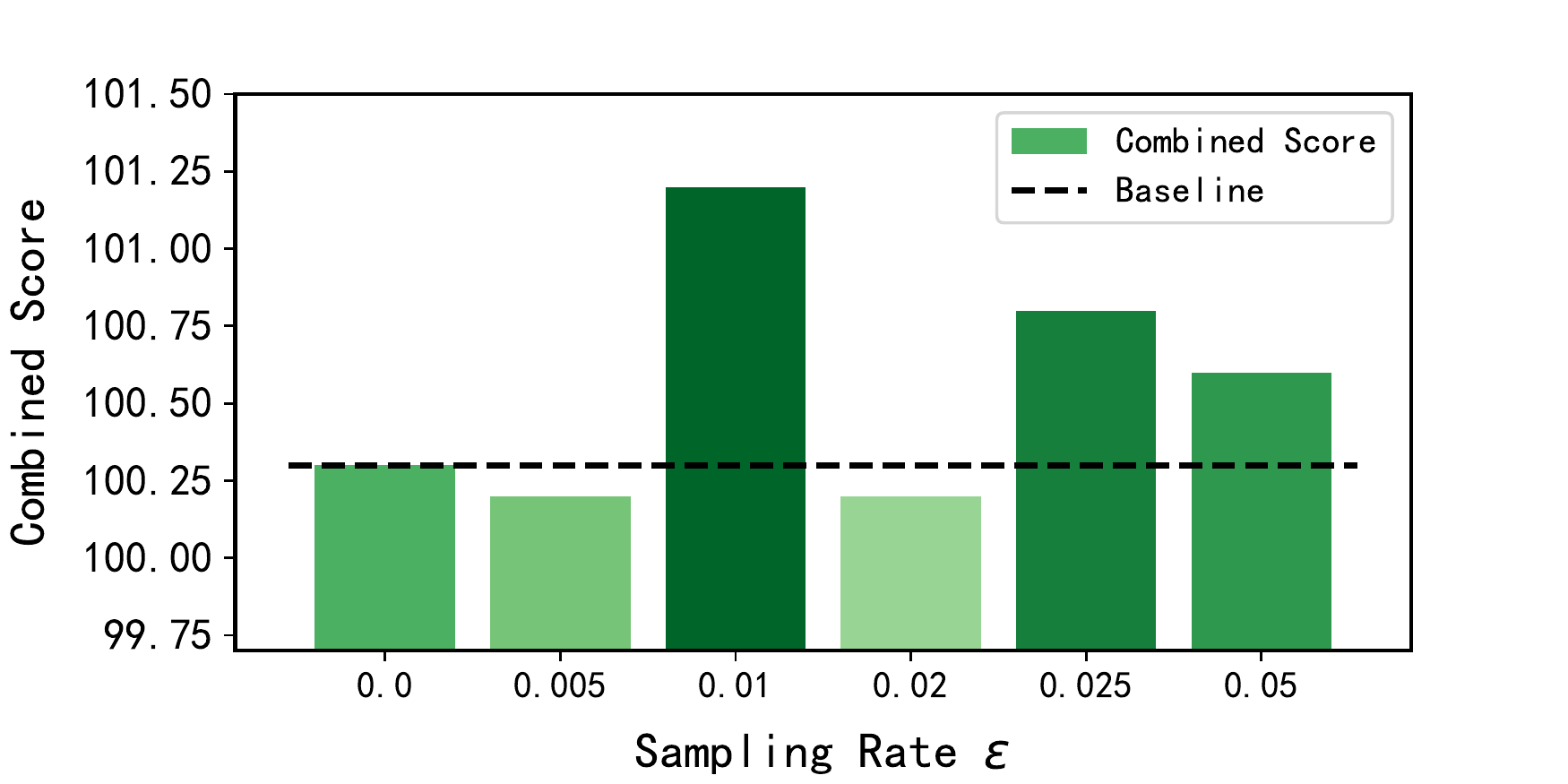}
    \caption{The combined score of different Sampling Rate $ \epsilon $}
    \label{fig:sampling_rate_score}
\end{figure}

Figure \ref{fig:sampling_rate_score} shows the effect of different sampling rate $ \epsilon $ of mixed training. With UBARv1 + R-Drop as the baseline, we explore $ \epsilon $ ranging from $ 0\% $ to $ 5\% $. 
When the sampling rate $ \epsilon = 1\% $, the combined score reaches the highest, exceeding the baseline. $ \epsilon $ of $ 2.5\% $ or $ 5\% $ can also lead to improvements. 
The sampling rate can hurt the system's performance if not appropriate, and whether too small or too large rate can lead to a decrease in the score compared to the baseline. 
Note that $ \epsilon = 1\% $ may seem small, but we believe that exposing the model to a small amount of data can make a difference, helping to mitigate the exposure bias. We provide more detailed results regarding the sampling rate in Appendix \ref{app:samplingrate}.

\subsection{Regularization Weight}
We discuss the effect of the weight $ \alpha $ in the KL-divergence regularization term of either R-Drop or R-Mask during the training of UBARv2. Here, we use UBARv1+SS as the baseline and add the KL-divergence regularization term to compare the scores of each metric with different coefficient weights.

\begin{figure}[h]
    \centering
    \includegraphics[width=0.52\textwidth]{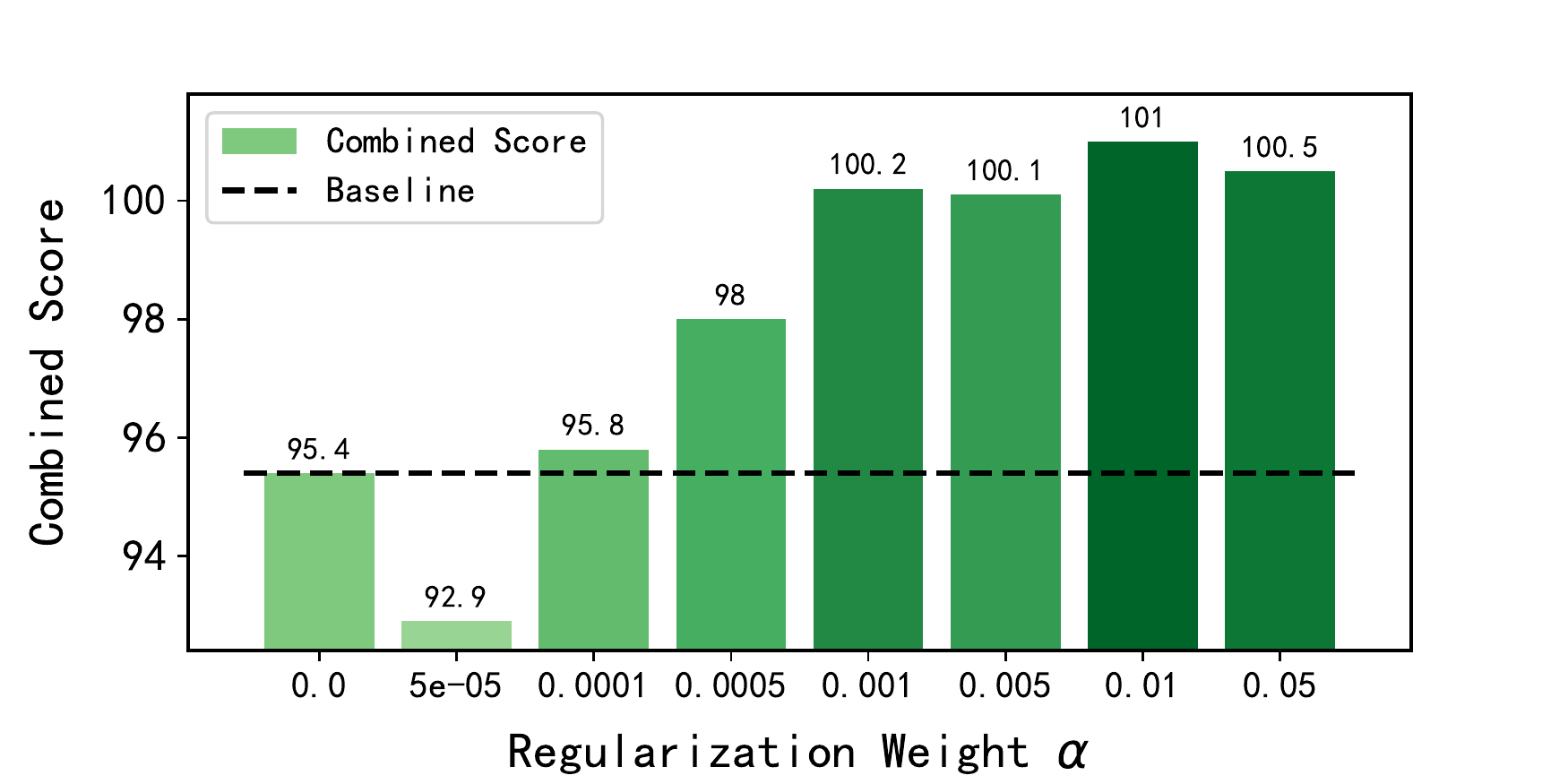}
    \caption{Regularization Weight $ \alpha $ from $ 0 $ to $ 0.05 $}
    \label{fig:rate_kl1_score_coarse}
\end{figure}

As shown in Figure \ref{fig:rate_kl1_score_coarse}, UBARv2 achieves the best results when $ \alpha $ is $ 0.01 $. 
We observe that all the different KL-divergence weights $\alpha$ except for 5e-5 provide performance gains. 
This is because the KL-divergence regularization term exists throughout the course of training , allowing the model to adaptively adjust itself. This also reflects the generalization ability of the R-Drop method, which can lead to a relatively stable boost.
We provide more detailed results regarding the regularization weight in Appendix \ref{app:regularization}.

\subsection{R-Mask Strategy}
The strategies for R-Mask tie closely with the strategies for session-level sampling as we have already identified belief state as the sampling target. R-Mask also requires a thorough discussion on how to construct the two sequences for the KL-divergence term.
Specifically, based on UBARv1 + R-Drop, we add a regularization term with R-Mask to UBARv2 at Stage 2 of mixed training.
We investigate the impact of different R-Mask strategies: (1) For the mask target, it can be either the sampled generated belief state or the ground-truth belief state. (2) For the mask position, there are two options: the two sequences for the regularization term are masked at the same positions or the two sequences are masked at different positions with the same mask rate. We search for suitable masking rates for different strategies.  
As shown in Table \ref{tab:rmask_strate}, the best strategy is to mask the ground-truth belief state at different positions for the two sequences at Stage 2, which offers more diversity to the two sequences and thus improves the model's generalization ability.
We provide results regarding the masking rate in Appendix  \ref{app:maskrate}

\begin{table}[ht]
    \centering
    \renewcommand{\arraystretch}{1.2}
    \begin{tabular}{c c | c}
        \toprule
        Mask Target & Mask Position  &  Combined \\
        \hline
        - & -  & 100.3\\
        Gen & Same &  100.1 \\
        Gen & Diff &  100.5 \\
        GT & Same &  100.0\\
        GT & Diff &  \textbf{101.6}\\
        \bottomrule
    \end{tabular}
    \caption{The combined scores with different R-Mask Strategies. Gen and GT denotes generated and ground-truth belief state respectively, Same means the masking the same positions for the two sequences and Diff means masking different positions.
    }
    \label{tab:rmask_strate}
\end{table}

\subsection{Case Study}
In this section, we present further discussions and empirical analyses of the effectiveness of the proposed method for mitigating the exposure bias in dialogs through case study.

As UBARv2 achieves a decent improvement on the combined score over UBARv1, it understandably has more correct cases than UBARv1. 
By looking at the cases in which UBARv1 predicted incorrectly but UBARv2 predicted correctly, we find that, in the majority of cases, UBARv1 just incorrectly predicts information like the belief state, while UBARv2 can get it right in the first place. 
Therefore, we are more concerned about whether UBARv2 can make the dialog context stay more consistent and coherent, and whether it can really bridge the gap between distributions in training and inference.

As shown in Table \ref{tab:case_correction2_new}, in the first turn, according to the ground truth, the user should be informed of the name and address of the hotel, but both models choose to ask for the hotel star rating to narrow down the choices. In the second turn, UBARv2 does a better job than UBARv1 at finding the missing hotel name in the context and providing it to the user in time. 
This case shows that UBARv1 still suffers from not being able to supplement entity names. While UBARv2 can supplement entity name appropriately, which reflects the fact that UBARv2 also can adaptively supplement and make amends in response to the current user utterance in order to stay consistent and coherent throughout the entire session and do it better than UBARv1.
However, it can be noted that the user also needs the address of the hotel, but UBARv2 does not supplement the address information. This indicates that UBARv2 can be further improved. 
\begin{table}[h]
    \centering
    \setlength\tabcolsep{2pt}
    \renewcommand{\arraystretch}{1.4}
    \scalebox{0.8}{
    \begin{tabular}{
        m{0.11\textwidth}<{\centering}|m{0.45\textwidth}
    }
    \toprule[2pt]
    
    User & i would like a moderate -ly priced hotel in the south, with free wifi. \\
    \hline
    GT Resp. & \mybox[yellow]{[value\_name]} is your choice. its located in \mybox[yellow]{[value\_address]} can i give you the address \\
    \hline
    UBARv1 & \textcolor{magenta}{i have [value\_choice] [value\_type] that meet your criteria. do you have a star rating preference?} \\
    \hline
    UBARv2 & \textcolor{magenta}{i have [value\_choice] [value\_type] that fit your criteria. do you have a preference for star rating?} \\
    \midrule
    \midrule
    
    User & you just gave me the address. does it have free parking? \\
    \hline
    GT Resp. & yes, both free internet and free parking. can i help you with anything else today? \\
    \hline
    UBARv1 & \textcolor{magenta}{yes, it does.} \\
    \hline
    UBARv2 & \textcolor{magenta}{\mybox[green]{[value\_name]} has free parking.} \\
    \bottomrule[2pt]
    \end{tabular}
    }
    
    \caption{Case Study: delexicalized responses generated by UBARv1 and UBARv2 of
    two consecutive dialog turns in dialog session PMUL0006 from MultiWOZ 2.0.
    }
    \label{tab:case_correction2_new}
\end{table}

We also want to address that current automatic evaluation metrics and static human evaluation are not adequate to quantitatively measure whether the exposure bias problem has been mitigated or not. There is a call for a more sophisticated and less labor-extensive evaluation. We provide two more case study in Appendix \ref{app:case}. 

\section{Related Work}
The architectures for end-to-end modeling of task-oriented systems can be coarsely divided into multi-decoder methods \cite{zhang2020task,zhang2020probabilistic,tseng2021transferable,wang2020multi,jeon2022dora,ramachandran2021causal} and pre-trained language models \cite{hosseini-asl2020a,peng2020soloist,kulhanek2021augpt,lin2020mintl,yang2021ubar,su2021multi,lee2021improving,sun2022bort,he2021galaxy}.
In terms of how to model the dialog context, session-level modeling has become popular with recent works \cite{yang2021ubar,su2021multi,lee2021improving,he2021galaxy}. 
Pre-training on relevant dialog corpus and multi-task learning are also employed to improve task completion.
SOLOIST \cite{peng2020soloist} is further pre-trained on a large dialog corpus with a multi-task objective. GALAXY \cite{he2021galaxy} is further pre-trained via semi-supervised learning which makes use of unlabeled dialog samples.
PPTOD \cite{su2021multi} proposes a multi-task pre-training strategy for dialogs with prompts. MTTOD \cite{lee2021improving} train a T5-based model with the auxiliary task.

The exposure bias problem is previously discussed and studied in the training process of neural machine translation \cite{bengio2015scheduled,ranzato2015sequence,shen2015minimum,wiseman-rush-2016-sequence,zhang2019bridging}. Contrastive learning is also used to reduce the exposure bias problem by learning in the representation space \cite{lee2020contrastive,liu2021simcls,pan2021contrastive}. 
\citet{wu-etal-2018-beyond} and \citet{wang2020exposure} shared some helpful insights on the relationship between exposure bias and error propagation. 
For TOD systems, exposure bias and error propagation exist in the multi-turn nature of dialogs. Some works have addressed the error propagation problem through data augmentation to increase the robustness of the systems \cite{zhang2020task,li2021retrieve,sun2022bort}. UBARv2 is the first work that designs methods for mitigating the exposure bias problem in task-oriented dialogs.


\section{Conclusion}
This work tries to mitigate the exposure bias problem in task-oriented dialog systems by proposing mixed training with session-level sampling and consistency regularization strategy R-Mask. UBARv2 achieves state-of-the-art performance on the end-to-end modeling task of MultiWOZ Evaluation, raising the combined score by over 1 point. By actively bridging the gap between training and inference, the model can stay more consistent and coherent with the generated context. 
We believe that the exposure bias problem exhibits in multi-turn dialogs is an interesting topic worth studying, and hope that UBARv2 can inspire future work to explore more methods to bridge the gap between training and inference for dialog systems.
\bibliography{anthology,custom}

\begin{thebibliography}{43}
\expandafter\ifx\csname natexlab\endcsname\relax\def\natexlab#1{#1}\fi

\bibitem[{Bengio et~al.(2015)Bengio, Vinyals, Jaitly, and
  Shazeer}]{bengio2015scheduled}
Samy Bengio, Oriol Vinyals, Navdeep Jaitly, and Noam Shazeer. 2015.
\newblock Scheduled sampling for sequence prediction with recurrent neural
  networks.
\newblock \emph{Advances in neural information processing systems}, 28.

\bibitem[{{Bengio} et~al.(2003){Bengio}, {Ducharme}, {Vincent}, and
  {Janvin}}]{bengio2003a}
Yoshua {Bengio}, Réjean {Ducharme}, Pascal {Vincent}, and Christian {Janvin}.
  2003.
\newblock A neural probabilistic language model.
\newblock \emph{Journal of Machine Learning Research}, 3(6):1137--1155.

\bibitem[{{Budzianowski} et~al.(2018){Budzianowski}, {Wen}, {Tseng},
  {Casanueva}, {Ultes}, {Ramadan}, and {Gasic}}]{budzianowski2018multiwoz}
Paweł {Budzianowski}, Tsung-Hsien {Wen}, Bo-Hsiang {Tseng}, Iñigo
  {Casanueva}, Stefan {Ultes}, Osman {Ramadan}, and Milica {Gasic}. 2018.
\newblock Multiwoz - a large-scale multi-domain wizard-of-oz dataset for
  task-oriented dialogue modelling.
\newblock In \emph{EMNLP 2018: 2018 Conference on Empirical Methods in Natural
  Language Processing}, pages 5016--5026.

\bibitem[{{Chen} et~al.(2019){Chen}, {Chen}, {Qin}, {Yan}, and
  {Wang}}]{chen2019semantically}
Wenhu {Chen}, Jianshu {Chen}, Pengda {Qin}, Xifeng {Yan}, and William~Yang
  {Wang}. 2019.
\newblock Semantically conditioned dialog response generation via hierarchical
  disentangled self-attention.
\newblock In \emph{ACL 2019 : The 57th Annual Meeting of the Association for
  Computational Linguistics}, pages 3696--3709.

\bibitem[{{Eric} et~al.(2019){Eric}, {Goel}, {Paul}, {Kumar}, {Sethi}, {Ku},
  {Goyal}, {Agarwal}, {Gao}, and {Hakkani-Tur}}]{eric2019multiwoz}
Mihail {Eric}, Rahul {Goel}, Shachi {Paul}, Adarsh {Kumar}, Abhishek {Sethi},
  Peter {Ku}, Anuj~Kumar {Goyal}, Sanchit {Agarwal}, Shuyang {Gao}, and Dilek
  {Hakkani-Tur}. 2019.
\newblock Multiwoz 2.1: A consolidated multi-domain dialogue dataset with state
  corrections and state tracking baselines.
\newblock In \emph{LREC}, pages 422--428.

\bibitem[{{Ham} et~al.(2020){Ham}, {Lee}, {Jang}, and {Kim}}]{ham2020end}
Donghoon {Ham}, Jeong-Gwan {Lee}, Youngsoo {Jang}, and Kee-Eung {Kim}. 2020.
\newblock End-to-end neural pipeline for goal-oriented dialogue systems using
  gpt-2.
\newblock In \emph{ACL 2020: 58th annual meeting of the Association for
  Computational Linguistics}, pages 583--592.

\bibitem[{He et~al.(2021)He, Dai, Zheng, Wu, Cao, Liu, Jiang, Yang, Huang, Si
  et~al.}]{he2021galaxy}
Wanwei He, Yinpei Dai, Yinhe Zheng, Yuchuan Wu, Zheng Cao, Dermot Liu, Peng
  Jiang, Min Yang, Fei Huang, Luo Si, et~al. 2021.
\newblock Galaxy: A generative pre-trained model for task-oriented dialog with
  semi-supervised learning and explicit policy injection.
\newblock \emph{arXiv preprint arXiv:2111.14592}.

\bibitem[{{Hosseini-Asl} et~al.(2020){Hosseini-Asl}, {McCann}, {Wu}, {Yavuz},
  and {Socher}}]{hosseini-asl2020a}
Ehsan {Hosseini-Asl}, Bryan {McCann}, Chien-Sheng {Wu}, Semih {Yavuz}, and
  Richard {Socher}. 2020.
\newblock A simple language model for task-oriented dialogue.
\newblock \emph{arXiv preprint arXiv:2005.00796}.

\bibitem[{Jeon and Lee(2022)}]{jeon2022dora}
Hyunmin Jeon and Gary~Geunbae Lee. 2022.
\newblock Dora: Towards policy optimization for task-oriented dialogue system
  with efficient context.
\newblock \emph{Computer Speech \& Language}, 72:101310.

\bibitem[{Kulh{\'a}nek et~al.(2021)Kulh{\'a}nek, Hude{\v{c}}ek, Nekvinda, and
  Du{\v{s}}ek}]{kulhanek2021augpt}
Jon{\'a}{\v{s}} Kulh{\'a}nek, Vojt{\v{e}}ch Hude{\v{c}}ek, Tom{\'a}{\v{s}}
  Nekvinda, and Ond{\v{r}}ej Du{\v{s}}ek. 2021.
\newblock Augpt: Auxiliary tasks and data augmentation for end-to-end dialogue
  with pre-trained language models.
\newblock \emph{arXiv preprint arXiv:2102.05126}.

\bibitem[{Le et~al.(2020)Le, Sahoo, Liu, Chen, and Hoi}]{le2020uniconv}
Hung Le, Doyen Sahoo, Chenghao Liu, Nancy Chen, and Steven~CH Hoi. 2020.
\newblock Uniconv: A unified conversational neural architecture for
  multi-domain task-oriented dialogues.
\newblock In \emph{Proceedings of the 2020 Conference on Empirical Methods in
  Natural Language Processing (EMNLP)}, pages 1860--1877.

\bibitem[{Lee et~al.(2020)Lee, Lee, and Hwang}]{lee2020contrastive}
Seanie Lee, Dong~Bok Lee, and Sung~Ju Hwang. 2020.
\newblock Contrastive learning with adversarial perturbations for conditional
  text generation.
\newblock \emph{arXiv preprint arXiv:2012.07280}.

\bibitem[{Lee(2021)}]{lee2021improving}
Yohan Lee. 2021.
\newblock Improving end-to-end task-oriented dialog system with a simple
  auxiliary task.
\newblock In \emph{Findings of the Association for Computational Linguistics:
  EMNLP 2021}, pages 1296--1303.

\bibitem[{Li et~al.(2021)Li, Yang, Quan, and Yu}]{li2021retrieve}
Yunhao Li, Yunyi Yang, Xiaojun Quan, and Jianxing Yu. 2021.
\newblock Retrieve \& memorize: Dialog policy learning with multi-action
  memory.
\newblock \emph{arXiv preprint arXiv:2106.02317}.

\bibitem[{Lin et~al.(2020)Lin, Madotto, Winata, and Fung}]{lin2020mintl}
Zhaojiang Lin, Andrea Madotto, Genta~Indra Winata, and Pascale Fung. 2020.
\newblock Mintl: Minimalist transfer learning for task-oriented dialogue
  systems.
\newblock \emph{arXiv preprint arXiv:2009.12005}.

\bibitem[{Liu and Liu(2021)}]{liu2021simcls}
Yixin Liu and Pengfei Liu. 2021.
\newblock Simcls: A simple framework for contrastive learning of abstractive
  summarization.
\newblock \emph{arXiv preprint arXiv:2106.01890}.

\bibitem[{Lubis et~al.(2020)Lubis, Geishauser, Heck, Lin, Moresi, van Niekerk,
  and Gasic}]{lubis2020lava}
Nurul Lubis, Christian Geishauser, Michael Heck, Hsien-Chin Lin, Marco Moresi,
  Carel van Niekerk, and Milica Gasic. 2020.
\newblock Lava: Latent action spaces via variational auto-encoding for dialogue
  policy optimization.
\newblock In \emph{Proceedings of the 28th International Conference on
  Computational Linguistics}, pages 465--479.

\bibitem[{{Mehri} et~al.(2019){Mehri}, {Srinivasan}, and
  {Eskenazi}}]{mehri2019structured}
Shikib {Mehri}, Tejas {Srinivasan}, and Maxine {Eskenazi}. 2019.
\newblock Structured fusion networks for dialog.
\newblock In \emph{Proceedings of the 20th Annual SIGdial Meeting on Discourse
  and Dialogue}, pages 165--177.

\bibitem[{Nekvinda and Du{\v{s}}ek(2021)}]{nekvinda2021shades}
Tom{\'a}{\v{s}} Nekvinda and Ond{\v{r}}ej Du{\v{s}}ek. 2021.
\newblock Shades of bleu, flavours of success: The case of multiwoz.
\newblock \emph{arXiv preprint arXiv:2106.05555}.

\bibitem[{Pan et~al.(2021)Pan, Wang, Wu, and Li}]{pan2021contrastive}
Xiao Pan, Mingxuan Wang, Liwei Wu, and Lei Li. 2021.
\newblock Contrastive learning for many-to-many multilingual neural machine
  translation.
\newblock \emph{arXiv preprint arXiv:2105.09501}.

\bibitem[{{Papineni} et~al.(2002){Papineni}, {Roukos}, {Ward}, and
  {Zhu}}]{papineni2002bleu}
Kishore {Papineni}, Salim {Roukos}, Todd {Ward}, and Wei-Jing {Zhu}. 2002.
\newblock Bleu: a method for automatic evaluation of machine translation.
\newblock In \emph{Proceedings of 40th Annual Meeting of the Association for
  Computational Linguistics}, pages 311--318.

\bibitem[{{Peng} et~al.(2020){Peng}, {Li}, {Li}, {Shayandeh}, {Liden}, and
  {Gao}}]{peng2020soloist}
Baolin {Peng}, Chunyuan {Li}, Jinchao {Li}, Shahin {Shayandeh}, Lars {Liden},
  and Jianfeng {Gao}. 2020.
\newblock Soloist: Few-shot task-oriented dialog with a single pre-trained
  auto-regressive model.
\newblock \emph{arXiv preprint arXiv:2005.05298}.

\bibitem[{Radford et~al.(2018)Radford, Narasimhan, Salimans, and
  Sutskever}]{radford2018improving}
Alec Radford, Karthik Narasimhan, Tim Salimans, and Ilya Sutskever. 2018.
\newblock Improving language understanding by generative pre-training.

\bibitem[{Ramachandran et~al.(2021)Ramachandran, Hashimoto, and
  Xiong}]{ramachandran2021causal}
Govardana~Sachithanandam Ramachandran, Kazuma Hashimoto, and Caiming Xiong.
  2021.
\newblock Causal-aware safe policy improvement for task-oriented dialogue.
\newblock \emph{arXiv preprint arXiv:2103.06370}.

\bibitem[{Ranzato et~al.(2015)Ranzato, Chopra, Auli, and
  Zaremba}]{ranzato2015sequence}
Marc'Aurelio Ranzato, Sumit Chopra, Michael Auli, and Wojciech Zaremba. 2015.
\newblock Sequence level training with recurrent neural networks.
\newblock \emph{arXiv preprint arXiv:1511.06732}.

\bibitem[{{Sanh} et~al.(2019){Sanh}, {Debut}, {Chaumond}, and
  {Wolf}}]{sanh2019distilbert}
Victor {Sanh}, Lysandre {Debut}, Julien {Chaumond}, and Thomas {Wolf}. 2019.
\newblock Distilbert, a distilled version of bert: smaller, faster, cheaper and
  lighter.
\newblock \emph{arXiv preprint arXiv:1910.01108}.

\bibitem[{Shen et~al.(2015)Shen, Cheng, He, He, Wu, Sun, and
  Liu}]{shen2015minimum}
Shiqi Shen, Yong Cheng, Zhongjun He, Wei He, Hua Wu, Maosong Sun, and Yang Liu.
  2015.
\newblock Minimum risk training for neural machine translation.
\newblock \emph{arXiv preprint arXiv:1512.02433}.

\bibitem[{Srivastava et~al.(2014)Srivastava, Hinton, Krizhevsky, Sutskever, and
  Salakhutdinov}]{srivastava2014dropout}
Nitish Srivastava, Geoffrey Hinton, Alex Krizhevsky, Ilya Sutskever, and Ruslan
  Salakhutdinov. 2014.
\newblock Dropout: a simple way to prevent neural networks from overfitting.
\newblock \emph{The journal of machine learning research}, 15(1):1929--1958.

\bibitem[{Su et~al.(2021)Su, Shu, Mansimov, Gupta, Cai, Lai, and
  Zhang}]{su2021multi}
Yixuan Su, Lei Shu, Elman Mansimov, Arshit Gupta, Deng Cai, Yi-An Lai, and
  Yi~Zhang. 2021.
\newblock Multi-task pre-training for plug-and-play task-oriented dialogue
  system.
\newblock \emph{arXiv preprint arXiv:2109.14739}.

\bibitem[{Sun et~al.(2022)Sun, Bao, Wu, and He}]{sun2022bort}
Haipeng Sun, Junwei Bao, Youzheng Wu, and Xiaodong He. 2022.
\newblock Bort: Back and denoising reconstruction for end-to-end task-oriented
  dialog.
\newblock \emph{arXiv preprint arXiv:2205.02471}.

\bibitem[{Tseng et~al.(2021)Tseng, Dai, Kreyssig, and
  Byrne}]{tseng2021transferable}
Bo-Hsiang Tseng, Yinpei Dai, Florian Kreyssig, and Bill Byrne. 2021.
\newblock Transferable dialogue systems and user simulators.
\newblock \emph{arXiv preprint arXiv:2107.11904}.

\bibitem[{Wang and Sennrich(2020)}]{wang2020exposure}
Chaojun Wang and Rico Sennrich. 2020.
\newblock On exposure bias, hallucination and domain shift in neural machine
  translation.
\newblock \emph{arXiv preprint arXiv:2005.03642}.

\bibitem[{Wang et~al.(2020)Wang, Zhang, Kim, and Gu}]{wang2020modelling}
Jianhong Wang, Yuan Zhang, Tae-Kyun Kim, and Yunjie Gu. 2020.
\newblock Modelling hierarchical structure between dialogue policy and natural
  language generator with option framework for task-oriented dialogue system.
\newblock \emph{arXiv preprint arXiv:2006.06814}.

\bibitem[{{Wang} et~al.(2020){Wang}, {Tian}, {Wang}, {Quan}, and
  {Yu}}]{wang2020multi}
Kai {Wang}, Junfeng {Tian}, Rui {Wang}, Xiaojun {Quan}, and Jianxing {Yu}.
  2020.
\newblock Multi-domain dialogue acts and response co-generation.
\newblock In \emph{ACL 2020: 58th annual meeting of the Association for
  Computational Linguistics}, pages 7125--7134.

\bibitem[{Wiseman and Rush(2016)}]{wiseman-rush-2016-sequence}
Sam Wiseman and Alexander~M. Rush. 2016.
\newblock \href {https://doi.org/10.18653/v1/D16-1137} {Sequence-to-sequence
  learning as beam-search optimization}.
\newblock In \emph{Proceedings of the 2016 Conference on Empirical Methods in
  Natural Language Processing}, pages 1296--1306, Austin, Texas. Association
  for Computational Linguistics.

\bibitem[{{Wolf} et~al.(2019){Wolf}, {Debut}, {Sanh}, {Chaumond}, {Delangue},
  {Moi}, {Cistac}, {Rault}, {Louf}, {Funtowicz}, and
  {Brew}}]{wolf2019huggingface}
Thomas {Wolf}, Lysandre {Debut}, Victor {Sanh}, Julien {Chaumond}, Clement
  {Delangue}, Anthony {Moi}, Pierric {Cistac}, Tim {Rault}, Rémi {Louf},
  Morgan {Funtowicz}, and Jamie {Brew}. 2019.
\newblock Huggingface's transformers: State-of-the-art natural language
  processing.
\newblock \emph{arXiv preprint arXiv:1910.03771}.

\bibitem[{Wu et~al.(2021)Wu, Li, Wang, Meng, Qin, Chen, Zhang, Liu
  et~al.}]{wu2021r}
Lijun Wu, Juntao Li, Yue Wang, Qi~Meng, Tao Qin, Wei Chen, Min Zhang, Tie-Yan
  Liu, et~al. 2021.
\newblock R-drop: regularized dropout for neural networks.
\newblock \emph{Advances in Neural Information Processing Systems}, 34.

\bibitem[{Wu et~al.(2018)Wu, Tan, He, Tian, Qin, Lai, and
  Liu}]{wu-etal-2018-beyond}
Lijun Wu, Xu~Tan, Di~He, Fei Tian, Tao Qin, Jianhuang Lai, and Tie-Yan Liu.
  2018.
\newblock \href {https://doi.org/10.18653/v1/D18-1396} {Beyond error
  propagation in neural machine translation: Characteristics of language also
  matter}.
\newblock In \emph{Proceedings of the 2018 Conference on Empirical Methods in
  Natural Language Processing}, pages 3602--3611, Brussels, Belgium.
  Association for Computational Linguistics.

\bibitem[{Yang et~al.(2021)Yang, Li, and Quan}]{yang2021ubar}
Yunyi Yang, Yunhao Li, and Xiaojun Quan. 2021.
\newblock Ubar: Towards fully end-to-end task-oriented dialog system with
  gpt-2.
\newblock In \emph{Proceedings of the AAAI Conference on Artificial
  Intelligence}, volume~35, pages 14230--14238.

\bibitem[{Zang et~al.(2020)Zang, Rastogi, Sunkara, Gupta, Zhang, and
  Chen}]{zang2020multiwoz}
Xiaoxue Zang, Abhinav Rastogi, Srinivas Sunkara, Raghav Gupta, Jianguo Zhang,
  and Jindong Chen. 2020.
\newblock Multiwoz 2.2: A dialogue dataset with additional annotation
  corrections and state tracking baselines.
\newblock \emph{arXiv preprint arXiv:2007.12720}.

\bibitem[{Zhang et~al.(2019)Zhang, Feng, Meng, You, and
  Liu}]{zhang2019bridging}
Wen Zhang, Yang Feng, Fandong Meng, Di~You, and Qun Liu. 2019.
\newblock Bridging the gap between training and inference for neural machine
  translation.
\newblock \emph{arXiv preprint arXiv:1906.02448}.

\bibitem[{Zhang et~al.(2020)Zhang, Ou, Wang, and Feng}]{zhang2020probabilistic}
Yichi Zhang, Zhijian Ou, Huixin Wang, and Junlan Feng. 2020.
\newblock A probabilistic end-to-end task-oriented dialog model with latent
  belief states towards semi-supervised learning.
\newblock \emph{arXiv preprint arXiv:2009.08115}.

\bibitem[{{Zhang} et~al.(2020){Zhang}, {Ou}, and {Yu}}]{zhang2020task}
Yichi {Zhang}, Zhijian {Ou}, and Zhou {Yu}. 2020.
\newblock Task-oriented dialog systems that consider multiple appropriate
  responses under the same context.
\newblock \emph{AAAI 2020 : The Thirty-Fourth AAAI Conference on Artificial
  Intelligence}, 34(5):9604--9611.

\end{thebibliography}
\bibliographystyle{acl_natbib}

\section{Appendix}
\label{sec:appendix}
%
\subsection{Results on Policy Optimization}
\label{app:policy}
Table \ref{tab:v2_policy_optimization} shows the results of UBARv2 in the policy optimization setting. Notice that UBARv2 did not achieve much improvement over UBARv1. This is because session-level sampling and R-Mask all target belief state, using the ground-truth belief state may render their advantages obsolete.

\begin{table*}[ht]
    \centering
    \setlength\tabcolsep{15pt}
    \renewcommand{\arraystretch}{1.2}
    \begin{tabular}{l | c c c c}
        \toprule
        Model & Inform & Success & BLEU & Comb \\
        \hline
        UniConv \cite{le2020uniconv} & 66.7 & 58.7 & 18.1 & 80.8 \\
        SFN \cite{mehri2019structured} & 93.4 & 82.3 & 14.1 & 101.9 \\
        HDSA \cite{chen2019semantically} & 87.9 & 79.4 & \textbf{20.7} & 104.4 \\
        LAVA \cite{lubis2020lava} & \textbf{95.9} & \textbf{93.5} & 10.8 & 105.5 \\
        HDNO \cite{wang2020modelling} & 93.3 & 83.4 & 17.8 & 106.1 \\
        MarCo \cite{wang2020multi} & 94.5 & 87.2 & 17.3 & \textbf{108.1} \\
        GALAXY \cite{he2021galaxy} & 92.8 &	83.5 & 19.9 & \textbf{108.1} \\
        \hline
        UBARV1 & 85.8 & 78.3 & 19.4 & 101.5 \\
        UBARv2 & 86.4 & 79.7 & 19.8 & 102.9 \\
        \bottomrule
    \end{tabular}
    \caption{
    Policy Optimization results on MultiWOZ Evaluation using ground-truth dialog states to generate responses.
    }
    \label{tab:v2_policy_optimization}
\end{table*}

\subsection{Domain Transfer}
\label{app:domain}
To examine the transfer ability of UBARv2 generalizing to unseen domains, we run zero-shot and few-shot experiments on the end-to-end modeling setting by excluding one domain out of the five domains that are available in validation and test set, and training UBARv2 on other four domains. Table \ref{tab:v2_domain_transfer} shows the results.
\begin{table*}[ht]
    \centering
    \setlength\tabcolsep{2.2pt}
    \renewcommand{\arraystretch}{1.2}
    \scalebox{0.85}{
    \begin{tabular}{c|c|c|c|c|c}
        \toprule
        \textbf{Evaluation on 4 Domains} & Except Hotel & Except Train & Except Attraction & Except Restaurant & Except Taxi \\
        \hline
        Base Model trained in-domain & \textbf{100.79} & 93.76 & 96.02 & 97.04 & 99.26 \\
        \hline
        Few-shot BM on new domain & \textbf{89.15} & 68.83 & 86.60 & 80.47 & 78.09 \\
        \hline
        UBARv2 on all domains & \textbf{106.81} & 99.38 & 100.08 & 100.58 & 101.75 \\
        \hline
        \hline
        \textbf{Evaluation on New Domain} & Hotel & Train & Attraction & Restaurant & Taxi \\
        \hline
        Zero-shot BM & 25.64 & 54.07 & 27.10 & 20.60 & \textbf{55.79} \\
        \hline
        Few-shot BM on new domain & 59.74 & 84.13 & 87.39 & 77.71 & \textbf{90.98} \\
        \hline
        UBARv2 on all domains & 92.04 & \textbf{102.27} & 102.04 & 101.21 & 97.51 \\

        \bottomrule
    \end{tabular}
    }
    \caption{
    Results of domain transfer. The first row is the base model of UBARv2 trained on the four domains and evaluated in-domain. The second row is the results of the base model fine-tuned with 100 new domain examples on the four domains. The last three rows are evaluations on the new domains with zero-shot or few-shot BM or UBARv2 trained on full data, respectively.
    }
    \label{tab:v2_domain_transfer}
\end{table*}

\subsection{Sampling Rate}
\label{app:samplingrate}
\begin{figure*}[h]
    \centering
    \includegraphics[width=1.05\textwidth]{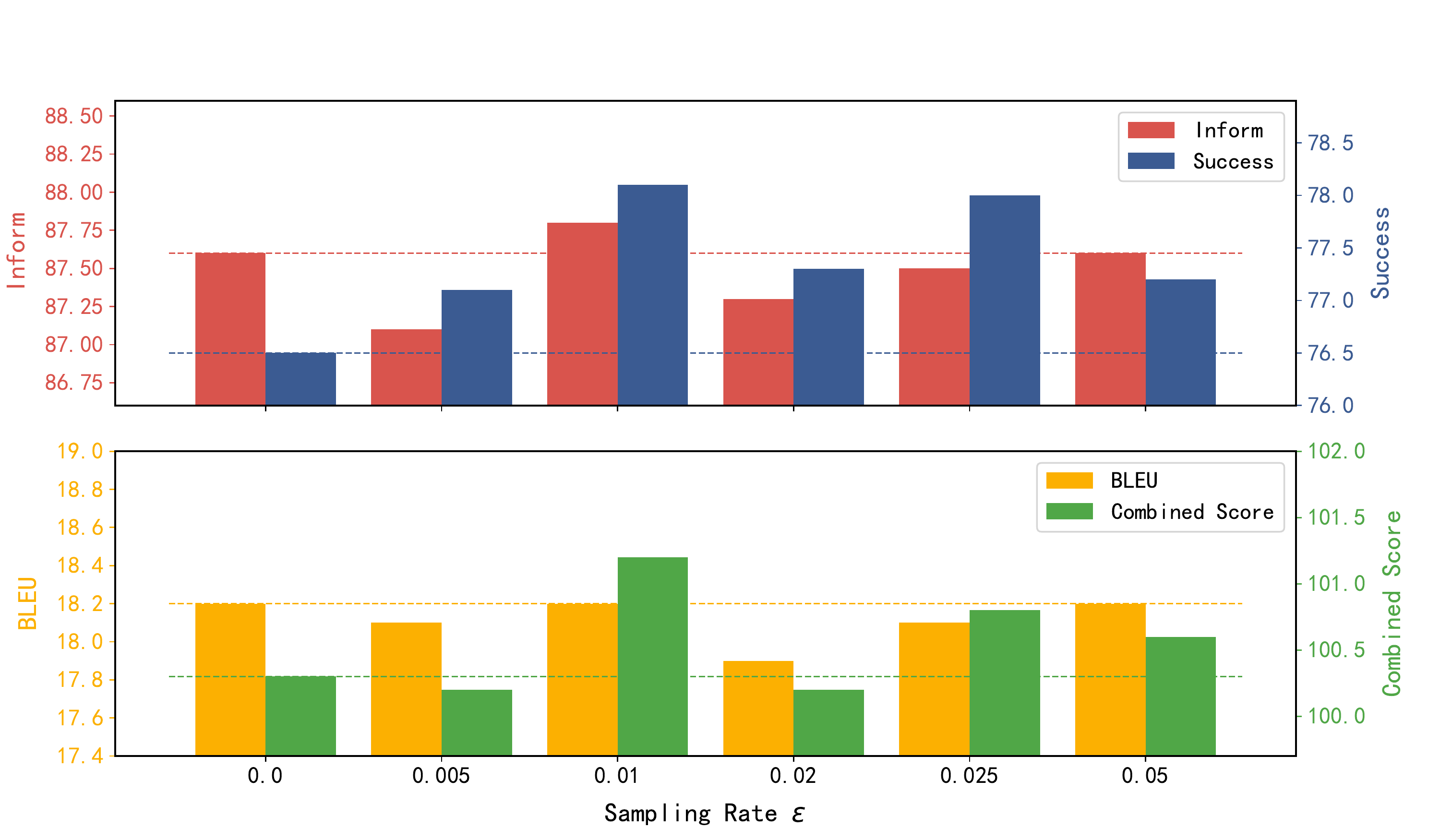}
    \caption{Sampling Rate $ \epsilon $}
    \label{fig:sampling_rate_all_metric}
\end{figure*}

As shown in Figure \ref{fig:sampling_rate_all_metric}, with UBARv1 + R-Drop as the baseline, the model completes the tasks better when the sampling rate is appropriate. When $ \epsilon $ is $ 0.01 $, it can maintain the fluency of responses.

\subsection{Effect on the Regularization weight}
\label{app:regularization}
\begin{figure*}[h]
    \centering
    \includegraphics[width=1.05\textwidth]{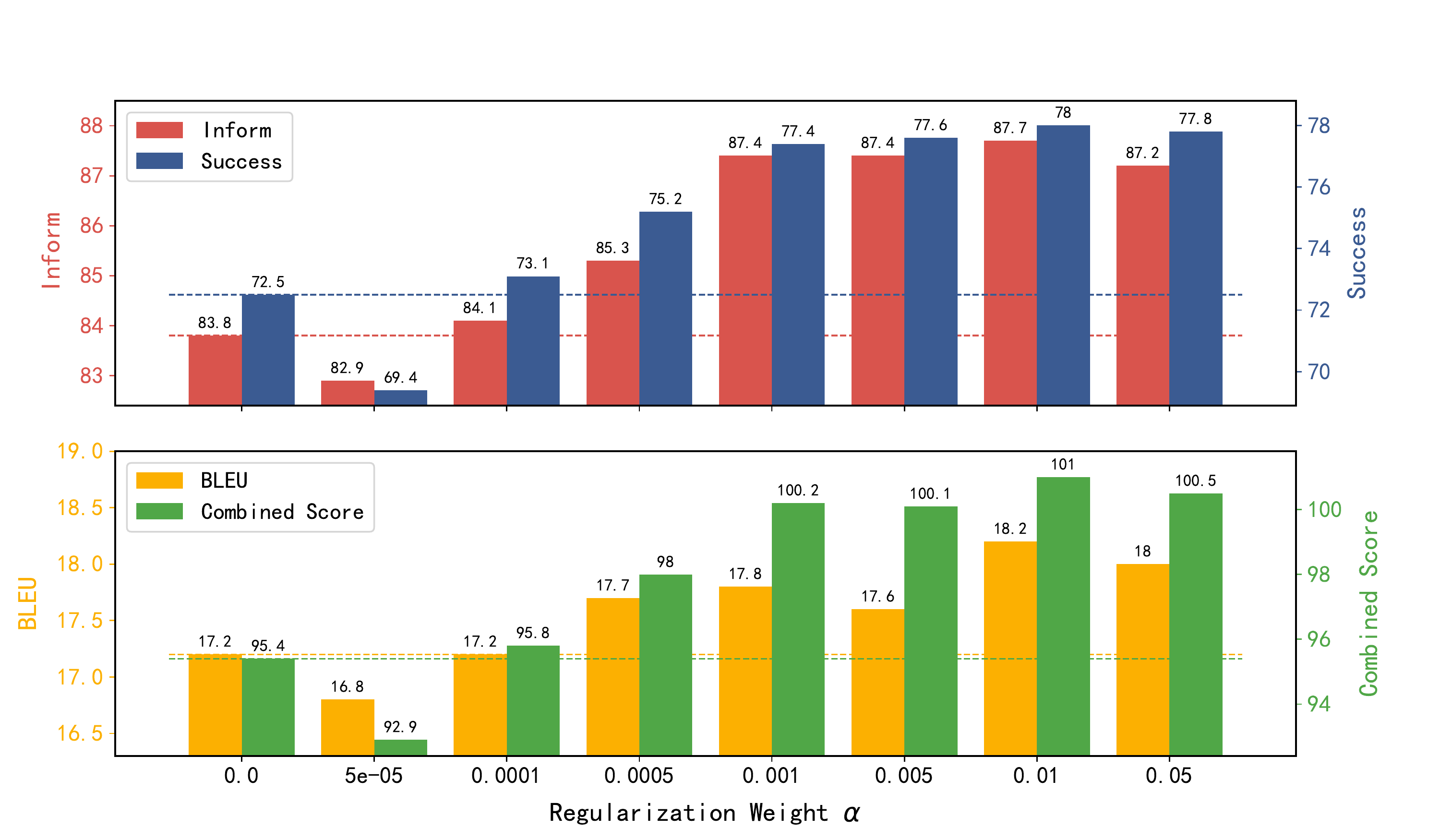}
    \caption{Regularization Weight $ \alpha $ ranging from $ 0 $ to $ 0.05 $}
    \label{fig:rate_kl1_all_metric_coarse}
\end{figure*}

\begin{figure*}[h]
    \centering
    \includegraphics[width=1.05\textwidth]{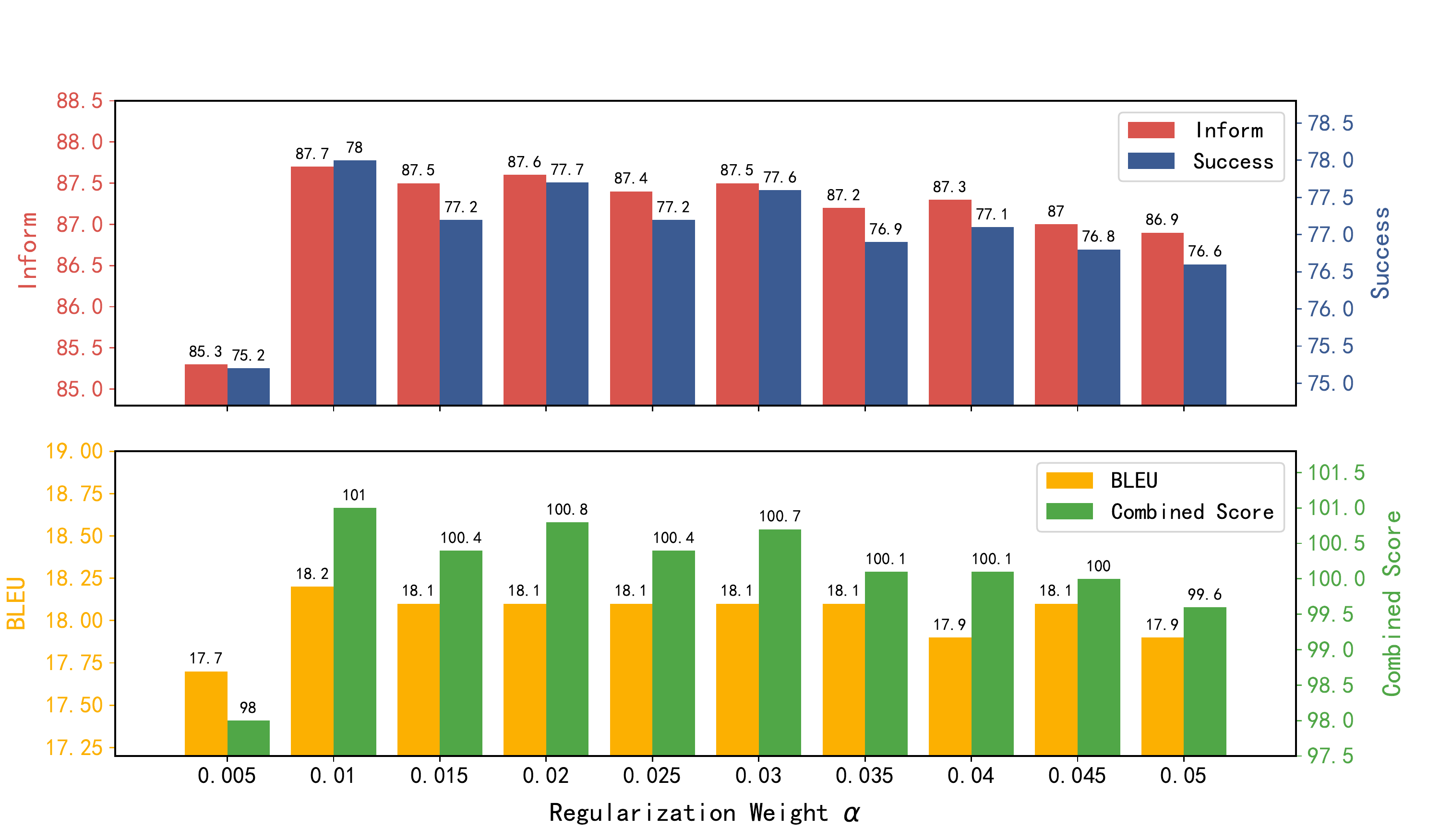}
    \caption{Regularization Weight $ \alpha $ ranging from $ 0.005 $ to $ 0.05 $}
    \label{fig:rate_kl1_all_metric_fine}
\end{figure*}

As shown in Figure \ref{fig:rate_kl1_all_metric_coarse}, among the results of all evaluation metrics corresponding to different regularization weight ranging from $ 0 $ to $ 0.05 $, the model achieves highest score when $ \alpha = 0.01 $. In order to find a better weight, we further explore it in a fine-grained setting from $ 0.005 $ to $ 0.05 $ in Figure \ref{fig:rate_kl1_all_metric_fine}, which shows that $ 0.01 $ is appropriate.

\subsection{Masking Rate}
\label{app:maskrate}
\begin{figure}[t]
    \centering
    \includegraphics[width=0.52\textwidth]{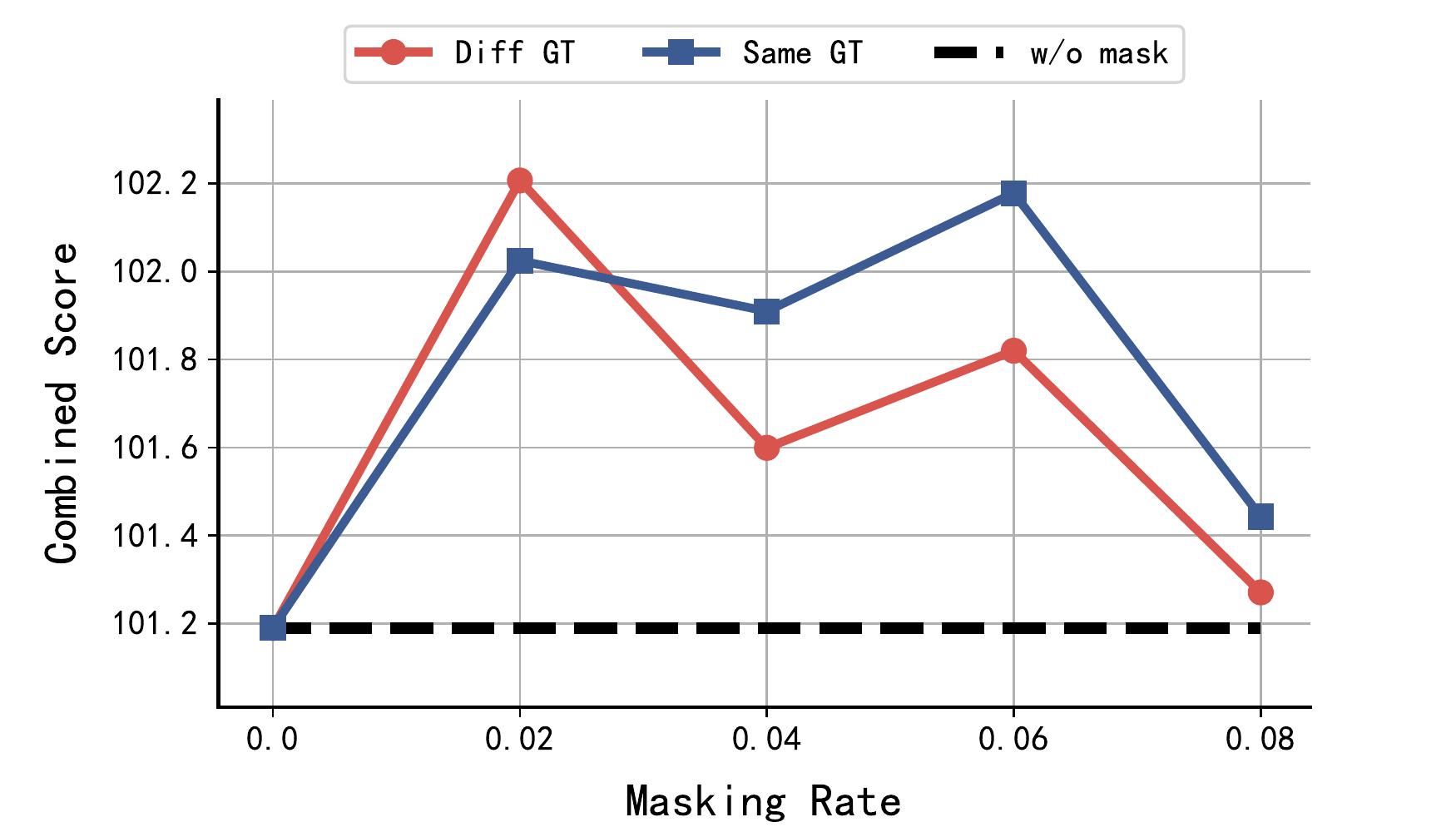}
    \caption{Masking Rate}
    \label{fig:mask_rate}
\end{figure}

Masking rate for ``Diff GT" and ``Same Gt" ranging from 0 to 0.08 is plotted in Figure \ref{fig:mask_rate}. 

\subsection{More Case Study}
\label{app:case}



As shown in Table \ref{tab:case_correction1_new}, where the user requests a recommendation for a modern European restaurant in downtown. In the third turn, the user should be given an explicit restaurant entity name according to the ground truth while UBARv1 and UBARv2 both choose to ask for the price of the restaurant to narrow down the choices. However, UBARv1 does not notice the context misses the necessary entity name and only simply provides the user with information such as an address, phone number, and price range in the fourth turn; on the contrary, UBARv2 can find logical inconsistency in context and provides key entity name in the fourth turn. From This case, we can see that UBAR using generated content as context does not completely avoid the problem of the missing entity name, and UBARv1 still has the error of not being able to supplement entity names. Instead, UBARv2 can supplement entity name appropriately, which reflects the fact that UBARv2 also can adaptively supplement and make amends in response to the current user utterance in order to stay consistent and coherent throughout the entire session and do it better than UBARv1. It is worth mentioning that at first we believe that the success of UBARv1 using all generated content comes from inconsistency between training and testing, i.e., the context that the model sees is not the ground truth but generated by the model itself, and are therefore concerned that removing exposure bias might cause UBARv2 to lose this helpful inconsistency, which means using mixed learning might cause the model to not learn to generate key entity words. Fortunately, this case eliminates our concerns and illustrates that the method used by UBARv2 to mitigate exposure bias in the dialog still retains and even improves the ability of the model to be consistent with the entire session.


It is difficult to determine whether the exposure bias is effectively mitigated and whether the difference between distributions in training and inference is bridged. Even work on machine translation and automatic summarization motivated by addressing exposure bias has also typically judged whether exposure bias is mitigated just based on the improvement in BLEU or ROUGE score. By this criterion, the improvement of UBARv2 is sufficient to prove that the motivation of the proposed method for mitigating the exposure bias in the dialog is reasonable, but we still want to find a case to show that UBARv2 can effectively mitigate exposure bias in the dialog. 

As shown in Table \ref{tab:case_gap_new}, UBARv1 using the original context can generate the key entity name in the current turn, but it can not respond correctly when using the generated context, which indicates that UBARv1 suffers from exposure bias in the dialog, i.e., UBARv1 has accumulated errors due to the generated context. However, again based on the generated context, UBARv2 still generates key entity names in the current turn, effectively mitigating the exposure bias.

\begin{table*}[h]
    \centering
    \setlength\tabcolsep{2pt}
    \renewcommand{\arraystretch}{1.4}
    \scalebox{0.8}{
    \begin{tabular}{
        m{0.11\textwidth}<{\centering}|m{0.80\textwidth}
    }
    \toprule[2pt]
    
    User & yes, please book 3 nights at the alpha - milton guest house for me. i would like to check in on wednesday. \\
    \hline
    GT Resp. & for how many people? \\
    \hline
    UBARv1 & how many people will be staying? \\
    \hline
    UBARv2 & your booking was successful. your reference number is [value\_reference]. is there anything else i can help you with? \\
    \midrule
    \midrule
    
    User & actually, i don't need a reservation. i am looking for a modern european restaurant in the center of town, though. \\
    \hline
    GT Resp. & the \mybox[yellow]{[value\_name]} is good and is located at the [value\_address] . can i reserve a table for you? \\
    \hline
    UBARv1 & \textcolor{magenta}{i have [value\_choice] options for you. do you have a price range in mind?} \\
    \hline
    UBARv2 & \textcolor{magenta}{there are [value\_choice] restaurant -s that meet your criteria. do you have a price range in mind?} \\
    \midrule
    \midrule
    
    User & can i get the address, phone number, and price range please? \\
    \hline
    GT Resp. & no problem. the address is [value\_address] . the phone number is [value\_phone] . it is in [value\_price] price range. \\
    \hline
    UBARv1 & \textcolor{magenta}{the address is [value\_address]. the phone number is [value\_phone]. the price range is [value\_price].} \\
    \hline
    UBARv2 & \mybox[green]{[value\_name]} is [value\_price] and located at [value\_address]. their phone number is [value\_phone]. \\

    \bottomrule[2pt]
    \end{tabular}
    }
    
    \caption{Case Study: delexicalized responses generated by UBARv1 and UBARv2 of
    three consecutive dialog turns in dialog session MUL1273 from MultiWOZ 2.0.
    }
    \label{tab:case_correction1_new}
\end{table*}

\begin{table*}[h]
    \centering
    \setlength\tabcolsep{2pt}
    \renewcommand{\arraystretch}{1.4}
    \scalebox{0.8}{
    \begin{tabular}{
        m{0.11\textwidth}<{\centering}|m{0.80\textwidth}
    }
    \toprule[2pt]
    
    User & not right now. i am also looking for an entertainment attraction on the east side of town. i 'll need their phone number and postcode. \\
    \hline
    GT Resp. & okay we have 2. \mybox[yellow]{[value\_name]}. \\
    \hline
    UBARv1 w/GT Context & \mybox[green]{[value\_name]} is located on postcode [value\_postcode] and the phone number is [value\_phone]. \\
    \hline
    UBARv1 & \textcolor{magenta}{sure! their phone number is [value\_phone] and their postcode is [value\_postcode].} \\
    \hline
    UBARv2 & \mybox[green]{[value\_name]} is located at postcode [value\_postcode]. their phone number is [value\_phone]. \\
    \midrule
    \midrule
    
    User & can you send me the phone number and postcode for the funky fun house? \\
    \hline
    GT Resp. & the phone number is [value\_phone] with post code [value\_postcode] \\
    \hline
    UBARv1 w/GT Context & the phone number is [value\_phone] and the postcode is [value\_postcode]. \\
    \hline
    UBARv1 & sure! the phone number is [value\_phone] and the postcode is [value\_postcode]. \\
    \hline
    UBARv2 & the phone number is [value\_phone] and the postcode is [value\_postcode]. is there anything else i can help you with? \\

    \bottomrule[2pt]
    \end{tabular}
    }
    
    \caption{Case Study: delexicalized responses generated by UBARv1 and UBARv2 of
    two consecutive dialog turns in dialog session PMUL3834 from MultiWOZ 2.0. UBARv1 w/GT Context is the response using annotated intermediate information.
    }
    \label{tab:case_gap_new}
\end{table*}


\end{document}